\documentclass[10pt,twocolumn,letterpaper]{article}

\usepackage{cvpr}
\usepackage{times}
\usepackage{epsfig}
\usepackage{graphicx}
\usepackage{amsmath}
\usepackage{amssymb}
\usepackage{multirow}
\usepackage{makecell}
\usepackage{caption}
\usepackage{geometry}
\usepackage{graphicx}
\usepackage{amsmath}
\usepackage{amsfonts}
\usepackage{booktabs}
\usepackage{mathrsfs}
\usepackage{amsbsy}
\usepackage{times}
\usepackage[ruled,vlined]{algorithm2e}
\usepackage[numbers]{natbib}
\newtheorem{definition}{$\mathbf{Definition}$} 
\newtheorem{theorem}{$\mathbf{Theorem}$}
\newtheorem{assumption}{$\mathbf{Assumption}$}
\usepackage{geometry}

\usepackage[breaklinks=true,bookmarks=false]{hyperref}

\cvprfinalcopy 

\def\cvprPaperID{****} 

\begin{document}

\title{\LARGE Reproduction of IVFS algorithm for high-dimensional topology preservation feature selection}
\author{Zihan Wang\\
	Department of Statistics and Data Science, Tsinghua University, Beijing 100084\\
	{\tt\small wangzh21@mails.tsinghua.edu.cn}
}

\maketitle

\begin{abstract}
Feature selection is a crucial technique for handling high-dimensional data. In unsupervised scenarios, many popular algorithms focus on preserving the original data structure. In this paper, we reproduce the IVFS algorithm introduced in AAAI 2020, which is inspired by the random subset method and preserves data similarity by maintaining topological structure. We systematically organize the mathematical foundations of IVFS and validate its effectiveness through numerical experiments similar to those in the original paper. The results demonstrate that IVFS outperforms SPEC and MCFS on most datasets, although issues with its convergence and stability persist.
\end{abstract}

\section{Introduction}
High-dimensional data is increasingly prevalent in a wide range of machine learning applications, including trading data analysis, natural language processing, and computer vision. However, the curse of dimensionality often results in computational inefficiencies and less effective models. Therefore, it is essential to select features that are both highly relevant to the target data and minimally redundant. This relevance can be measured using metrics such as KNN accuracy, distances between persistent diagrams, and other related methods.

Feature selection methods \citep{Guyon2003} aim to reduce the number of features while maintaining or even improving the performance of supervised classifiers compared to using the entire feature set \citep{SOLORIOFERNANDEZ2020}. Additionally, feature selection helps identify relevant and redundant features, reduces storage and processing time, and mitigates the curse of dimensionality \citep{BOLONCANEDO2019}. \citet{Li2020} proposed an unsupervised feature selection algorithm that effectively preserves pairwise distances and the topological structure of the full dataset. They highlighted two significant challenges in existing similarity-preserving feature selection methods: high dimensionality and large sample sizes. Their proposed algorithm addresses analytically intractable problems and is efficient for large-scale datasets, as demonstrated through extensive experimentation. In this paper, we implement the algorithm using our own code\footnote{\url{https://github.com/Wang-ZH-Stat/IVFS.git}} and reproduce the experimental results to compare and evaluate the performance of three algorithms. 

The paper is organized as follows: In Section \ref{description}, we outline the problems to be addressed and introduce key concepts such as inclusion value, computational topology, and distances between persistent diagrams. In Section \ref{solution}, we reorganize the solution approach from the original paper, focusing on its mathematical foundations and implementation, and detail the methods for comparison and evaluation metrics in the numerical experiments. In Section \ref{experiments}, we present the main results of our numerical experiments, conducted with our code, and discuss the outcomes in terms of robustness, stability, and the efficiency-capacity trade-off. In Section \ref{conclusions}, we provide our conclusions.

\section{Problem setup}
\label{description}
The data matrix is denoted by $X\in\mathbb{R}^{n\times d}$ with $n$ samples and $d$ covariates. Our goal is to find the best $d_0 < d$ features according to some criteria, associated with a loss function $\mathcal{L}$. Denote $\mathcal{I} = \{1, 2, ..., d\}$ the indices of all features, and $\mathcal{M}_{\tilde{d}}=\{\sigma\in \mathcal{I}^{\tilde{d}}:\sigma_i \neq\sigma_j, \forall i\neq j\}$the set of all size-$\tilde{d}$ subsets of $\mathcal{I}$. The goal is to minimize the objective function
$$
\mathop{\min}_{\mathcal{F}\in \mathcal{M}_{d_0} }\mathcal{L}_{\mathcal{F}}\triangleq\mathop{\min}_{\mathcal{F}\in \mathcal{M}_{d_0} }\mathcal{L}(X_{\mathcal{F}}).
$$
Next, we will give some important definitions to enhance the understanding of the algorithm later.

\subsection{Inclusion value}
The IVFS algorithm relies on repeatedly sampling random subset $\tilde{\mathcal{F}}$ of arbitrarily $\tilde{d}$ features (not necessarily equal to $d_0$), equipped with a subset score function $s(\tilde{\mathcal{F}}; X):\mathbb{R}^{\tilde{d}}\xrightarrow{}\mathbb{R}^{\tilde{d}}$ which assigns score to each selected feature by evaluating the chosen random subset. In principle, a high score should correspond to a small loss. The individual feature score, which serves as the filter of the unified selection scheme, essentially depends on the subset score function $s(\cdot)$ defined above.
\begin{definition}
	Suppose $1\le \tilde{d}\le d$. The inclusion value of feature $f\in \mathcal{I}$ at dimension $\tilde{d}$ associated with $s(\cdot)$ is 
	$$
	IV_{\tilde{d}}(f)=\frac{\sum_{\sigma\in \mathcal{M}_{\tilde{d}}^f}s_{\sigma}(f)}{\tbinom{d-1}{\tilde{d}-1}},
	$$
	where $\mathcal{M}_{\tilde{d}}^f=\{\sigma\in\mathcal{M}_{\tilde{d}}:f\in\sigma\}$ is the collection of subsets with size $\tilde{d}$ that contains feature $f$, and $s_{\sigma}(f)$ is the score assigned to feature $f$ by computing $s(\sigma;X)$.
\end{definition}

Intuitively, the inclusion value illustrates how much gain in score a feature $f$ could provide on average, when it is included in the feature subset of size $\tilde{d}$. Inclusion value is one of the core concepts of IVFS algorithm. Its definition is very simple, but when the data dimension is high, it is not feasible to calculate the accurate value.

\subsection{Computational topology}
In this section, we provide some intuition to several important concepts in computational topology \citep{2017Persistence}. In Euclidean space, the most commonly used complex is the Vietoris--Rips complex. It is formed by connecting points with distance smaller than a given threshold $\alpha$. If we gradually increase $\alpha$ from 0 to $\infty$, the number of edges will increase from 0 to $n^2$ eventually. The distance associated with each edge, is called the filtration for Rips complex. As $\alpha$ increases, topological features with different dimension will appear and disappear. We call the pair of birth and death time of a $p$-dimensional topological feature as a $p$-dimensional persistent barcode. The $p$-dimensional persistent diagram is a multiset of all these barcodes. Note that we can always normalize the filtration function to be bounded in $[0, 1]$. Often, barcodes with length less than a small number $\epsilon$ are regarded as noise and eliminated from the diagram. 

In this study, we focus on Rips complex which is widely used for real-valued data. The filtration of Rips complex is based on distances between data points. We will mainly focus on Euclidean distance. Consider a distance matrix $(D_{ij})$, with the $(i, j)$-th entry defined as $D_{ij} = \Vert x_i-x_j\Vert_2$, where $x_i$ and $x_j$ are two sample points, and $\Vert\cdot\Vert_2$ is the $l_2$ norm for vectors. We divide $D$ by its largest entry to normalize all distances to [0, 1].

\subsection{Distances between persistent diagrams}
\label{subsec.dis}
Finally, we define the distances between persistent diagrams \citep{2009Computational}. The following two distances measures between diagrams are widely used in topological data analysis (TDA). 

A persistence diagram is a multiset of points in the extended plane, $R^2$. Let $X$ and $Y$ be two persistence diagrams. To define the distance between them, we consider bijection $\eta : X \to Y $ and record the supremum of the distances between corresponding points for each. We then measure the distance between points $x,y$ with $L_{\infty}$ norm and taking the infimum over all bijections, and we define the formula below as bottleneck distance between the diagrams: 
$$
W_{\infty}(X,Y)=\inf_{\eta:X \to Y}\sum_{x \in X}||x-\eta(x)||_{\infty}.
$$
This is Bottleneck distance.

A drawback of the bottleneck distance is its insensitivity to details of the bijection beyond the furthest pair of corresponding points. To remedy this shortcoming, we introduce the degree-$q$ Wasserstein distance between diagrams $X$ and $Y$ for any $q>0$. It takes the sum of $q$-th powers of the $L_{\infty}$ distances between corresponding points and  minimizing over all bijections: 
$$
W_q(X,Y)=\left\{\inf_{\eta:X \to Y}\sum_{x \in X}||x-\eta(x)||_{\infty}^q\right\}^{1/q}.
$$
From the definition above we can see that both Bottleneck distance and Wasserstein distance do not have explicit solutions. To estimate their upper-bound, we have the theorems.

\begin{theorem}
\label{the:1}
	(Stability for filtrations) Let $K$ be a simplicial complex and $f,g : K \to R$ are two monotonic functions. For each dimension $p$, the bottleneck distance between $X,Y$, which denote the persistent diagrams built upon $K$ using filtration functions $f$ and $g$, is bounded from above by the $L_{\infty}$-distance between the functions, that is:
    $$
    W_{\infty}(X, Y ) \le ||f - g||_{\infty}.
    $$
\end{theorem}
\begin{theorem}
\label{the:2}
 	(Stability for Lipschitz functions) Let $f, g : K \to R$ being tame Lipschitz functions on a metric space whose triangulations grow in polynomial order with constant exponent $j$. Then there are constants $C$ and $k > j, k \ge 1$ such that the degree-$q$ Wasserstein distance between the diagrams $X,Y$ has 
 	$$
        W_q(X,Y)\le C \Vert f - g\Vert^{1-k/q},\quad\forall q \ge k.
        $$
\end{theorem}

\section{Solution}
In this section, we give the solution of feature selection problem by topology preservation. We will review the original paper's idea completely.

\label{solution}
\subsection{Topology preservation}
Many embedded methods are associated with clustering or nearest neighbor graph embedding, which in some sense address more on the local manifold structure. In this paper, we look at similarity preservation problem from a new view: topology. The merit of TDA comes from its capability to encode all the topological information of a point set $X$ by persistent diagram, denoted as $D(X)$ herein. In Euclidean space, $D(X)$ is tightly related to pairwise sample distances. This motivates us to consider similarity preservation from a TDA perspective. If we hope to select covariates $X_{\mathcal{F}}$ to preserve the topological information of the original data, the persistent diagram $D(X_{\mathcal{F}})$ generated by $X_{\mathcal{F}}$ should be close to the original diagram $D(X)$. More precisely, TDA offers a new space (of persistent diagrams) in which we compare and preserve the distances. We refer this property as ``topology preservation'', which is very important especially when applying TDA after feature selection.

Now we are ready to formally state the objective function. Recall the notations $\mathcal{I}=\{1,2,\dots,d\}$, and $\mathcal{M}_{\tilde{d}}=\{\sigma\in \mathcal{I}^{\tilde{d}}:\sigma_i\neq \sigma_j, \forall i\neq j\}$. To achieve topology preserving feature selection, we minimize following loss function
\begin{equation}
	\mathop{\min}_{\mathcal{F}\in \mathcal{M}_{d_0}}\omega_{*}(D(X), D(X_\mathcal{F})),
	\label{con:1}
\end{equation}
where $\omega_{*}$ denotes Wasserstein or bottleneck distance. In this way, we find the subset that best preserves the topological signatures of original data. However, the mapping between feature space and persistent diagram is so sophisticated that analytical approach to (\ref{con:1}) is hard to derive. This is exactly the circumstance where IVFS is particularly effective.

In Section~\ref{subsec.dis}, we have defined the distances between two persistent diagrams and the stability Theorem~\ref{the:1} and Theorem~\ref{the:2}. The theorems say that when we change filtration from $f$ to $g$, the change in persistent diagrams would be bounded linearly in the $\Vert f-g\Vert_{\infty}$ for Bottleneck distance, and polynomially for Wasserstein distance. Since the filtration for Rips complex is the pairwise distances, we can alternatively control the $l_{\infty}$ norm of the difference between two distance matrices. Thus, we substitute our objective to
\begin{equation}
\label{con:2}
	\mathop{\min}_{\mathcal{F}\in \mathcal{M}_{d_0}}\Vert D-D_{\mathcal{M}}\Vert_{\infty},
\end{equation}
where $\Vert A\Vert_{\infty}=\max_{i,j}A_{i,j}$, $D$ and $D_{\mathcal{F}}$ are the distance matrix before and after feature selection. By (\ref{con:2}), we get rid of the expensive computation of persistent diagrams, making the algorithm applicable to real world applications. Nevertheless, optimization regarding $l_{\infty}$ is still non-trivial.

\subsection{IVFS algorithm}
IVFS algorithm is constructed based on inclusion value estimation, as summarized in Algorithm~\ref{ivfs}. Roughly speaking, the algorithm selects features with highest estimated inclusion value, which is derived via $k$ random sub-samplings of both features and observations. One benefit is that IVFS considers complicated feature interactions by evaluating subset of features together in each iteration.

\begin{algorithm}[t]
	\SetAlgoLined
	\textbf{Input}: Data matrix $X\in \mathbb{R}^{n\times d}$; the number of subsets $k$; the number of features used for each subset $\widetilde{d}$; the number of samples for each subset $\widetilde{n}$; target dimension $d_0$ \\
	\textbf{Output}: Select top $d_0 $ features with highest score \\
	\textbf{Initialize}: Counters for each feature $c_i=0,i=1,...,d$; Cumulative score for each feature $S_i = 0$ \\
	\For {$t=1$ to $k$} {
		Randomly sample a size $\widetilde{d}$ feature set $\mathcal{F}\in \mathcal{M}_{\widetilde{d}}$ \\
		Randomly sub-sample $X_{\mathcal{F}}^{sub} \in \mathbb{R}^{\widetilde{n} \times \widetilde{d}}$, with $\widetilde{n}$ observations and features in $\mathcal{F}$\\
		\For {$f$ in $\mathcal{F}$} {
			Update counter $c_f=c_f+1$\\
			Update score $ S_f = S_f + s_{\mathcal{F}}(f)$ 
		}
	}
	Set $S_i=S_i/c_i$ for $i=1, 2, ..., d$
	\caption{IVFS}
	\label{ivfs}
\end{algorithm}

The IVFS scheme (Algorithm~\ref{ivfs}) can be directly adapted to the topology preservation problem (\ref{con:2}). We substitute score function as
$$
s_{\mathcal{F}}(f)=-\Vert D_{\mathcal{F}}-D\Vert_{\infty},\quad \forall f\in \mathcal{F},
$$
and all other steps remain the same. This is called the IVFS-$l_{\infty}$ algorithm. We can easily extend this algorithm to other reasonable loss functions. Denote $\Vert D-D_{\mathcal{F}}\Vert_1=\sum_{i,j}|D_{ij}-D_{\mathcal{F}_{ij}}|$ and $\Vert D-D_{\mathcal{F}}\Vert_2$ the matrix Frobenius norm. One should expect that minimizing these two norms of $(D - D_{\mathcal{F}})$ to be good alternatives, because small $\Vert D-D_{\mathcal{F}}\Vert_1$ or $\Vert D-D_{\mathcal{F}}\Vert_2$ is likely to result in a small $\Vert D_{\mathcal{F}}-D\Vert_{\infty}$ as well. We will call these two methods IVFS-$l_1$ and IVFS-$l_2$ algorithm respectively. Then, we will analyze the convergence and correctness of IVFS algorithm. The following two theorems are proved in the original paper. 

\begin{theorem}
    \label{the:3}
    (Asymptotic $k$) Suppose $k \to \infty, \tilde{n} = n, s_{\sigma}(f)$ has finite variance for any $f \in \mathcal{I}$, and the $IV_{d_0}$ for different features are all distinct, then IVFS algorithm will select top $d_0$ features with highest $IV_{\tilde{d}}$ with probability $1$.
\end{theorem}

Theorem \ref{the:3} ensures the convergence of IVFS algorithm. Further more, by central theorem, when $k \to \infty$ we have for some $\tau$ and $\forall f$,
$$
\sqrt{k}\{\hat{IV}_{\tilde{d}}(f)-IV_{\tilde{d}}(f)\}\to_d N(0,\tau^2).
$$

\begin{assumption}
(Monotonicity) There exists a $d$-dimensional set $\Omega \in \mathcal{M}_{\tilde{d}}$ such that $\forall \mathcal{F} , \mathcal{F}' \in \mathcal{M}_{\tilde{d}}$, if $(\mathcal{F} \cap \Omega) \subseteq (\mathcal{F}' \cap \Omega)$, then $s(\mathcal{F})\le s(\mathcal{F}')$.
\label{ass:1}
\end{assumption}

\begin{theorem}
\label{the:4}
 	(Optimality) Under Assumption~\ref{ass:1}, we have
 	$$
        \Omega = \arg \max_{\mathcal{F} \in \mathcal{M}_{\tilde{d}}}s(\mathcal{F}),
        $$
 	and $\Omega$ is the set of $\tilde{d}$ features with the highest $IV_{\tilde{d}}$.
\end{theorem}

Theorem \ref{the:4} ensures the correctness of IVFS algorithm.

\subsection{Methods and tuning parameters}
We compare several popular similarity preserving methods:
\begin{itemize}
    \item $\mathbf{SPEC}$: We use the second type algorithm which performs the best.
    \item $\mathbf{MCFS}$: We run MCFS with the number of clusters $M = \{5,\ 10,\ 20,\ 30\}$.
    \item $\mathbf{IVFS-l_{\infty}}$ and variants: We try following combinations: $\tilde{d}=\{0.1\ 0.3\ 0.5\}\times d$, $\tilde{n}=\{0.1\ 0.3\ 0.5\}\times n$ (if sample size of dataset is not more than 200) or $\{100\}$ (otherwise). We run experiments with $k$ = 1000, 3000, 5000. It is worth noting that our parameter combination here is slightly different from that of the original paper. This is because that our computing power of computer equipment is not as strong as that of Baidu Research.
\end{itemize}

Here we don't compare GLSPFS algorithm like that in the original paper, since we don't find the open source GLSPFS algorithm in the python language. This has little effect on the whole experiment, because according to the results of the original paper, it is enough to compare with the other two algorithms.

\subsection{Evaluation}
In Section~\ref{experiments}, we will compare method by various widely adopted metrics that can well evaluate the quality of selected feature set. Their details are as follows.
\begin{itemize}
    \item \textbf{KNN accuracy}. Following\citep{Zheng2013On}, etc., we test local structure preservation by KNN classification. Each dataset is randomly split into 80\% training sets and 20\% test set, on which the test accuracy is computed. We repeat the procedure 10 times and take the average accuracy. For the number of neighbors, we adopt $K\in\{1, 3, 5, 10\}$ and report the highest mean accuracy.
    
    \item \textbf{K-means accuracy}. K-means is an unsupervised clustering algorithm. Given the true labels, we can calculate the accuracy of clustering. We use it because it is from the documentation of \textbf{skfeature} module. This metric was not considered in the original paper. Here we make a supplement.
    
    \item \textbf{Normalized Mutual Information}. NMI \citep{2009Normalized} is often used in clustering to measure the similarity of two clustering results. NMI is an important measure of community detection, which can objectively evaluate the accuracy of a community partition compared with the standard partition. The range of NMI is 0 to 1. The higher the NMI, the more accurate the division. This metric was not considered in the original paper. Here we make a supplement.

    \item \textbf{Distances between persistent diagrams}. For $X$ and $X_{\mathcal{F}}$, we compute the 1-dimensional persistent diagram $D$ and $D_F$ for $X$ and $X_{\mathcal{F}}$ with $\alpha = 0.8$ and drop all barcodes with existing time less than $\epsilon=0.1$. Wasserstein ($\omega_1^1$) and Bottleneck ($\omega_{\infty}$) distances are computed between the diagrams. It is worth noting that here we choose a different $\alpha$ and $\epsilon$ from the original paper. This is because we don't know what third-party package the original authors used to calculate the distances. Here we choose the parameters that are suitable for our method.

    \item \textbf{Norms between distance matrix}. In real world sometimes the purpose of feature selection is to preserve the sample distance, and a straightforward way to evaluate different algorithms’ sample distance preserving ability is to measure the change between distance matrices $D$ and $D_F$. To make the measurement more convincing, we compute $\Vert D-D_F \Vert_1$, $\Vert D-D_F \Vert_2$ and $\Vert D-D_F \Vert_{\infty}$ to evaluate the closeness between $D$ and $D_F$.

    \item \textbf{Running time}. We compare the running time for a single run, with fixed parameter setting as same as original authors: \textbf{MCFS}: k = 10 and \textbf{IVFS}: $k=1000, \tilde{n}=0.1n, \tilde{d}=0.3d$.
\end{itemize}

\section{Experiments}
\label{experiments}
Following the original paper, we carry out extensive experiments on popular high-dimensional datasets from ASU feature selection database\citep{2016Feature}. The summary statistics are provided in Table~\ref{tab:1}. For all datasets, the features are standardized to mean zero and unit variance, and 300 features are selected. Different from the original paper, we use three additional datasets, including \emph{Arcene}, \emph{ORL} and \emph{Yale}. In this way, we can more comprehensively evaluate the effectiveness of IVFS algorithm, similar to the two new evaluation criteria as mentioned earlier.

\subsection{Results}
Table~\ref{tab:1} (on the last page of this paper) summarizes the results. For algorithms with tuning parameters, we report the best result among all parameters and the number of selected features for each metric. From Table~\ref{tab:1}, most of our results are close to the original paper. This shows that we have well reproduced the IVFS algorithm and evaluation criteria in the original paper.

In Table~\ref{tab:1}, the bold numbers represent the optimal values of given evaluation criteria in given dataset. We observe IVFS-$l_{\infty}$ provides smallest $\omega_1^1$, $\omega_{\infty}$, $L_{\infty}$, $L_1$ and $L_2$ on almost all datasets. Thus, the distance and topology preserving capability is essentially improved. Moreover, IVFS-$l_{\infty}$ beats other methods in terms of KNN accuracy, K-means accuracy and NMI, which indicates its superiority on supervised tasks and local manifold preservation. The discussion of running time will come later.

\subsection{Robustness}
\begin{figure}[ht]
	\begin{center}
		\includegraphics[width=0.87\linewidth]{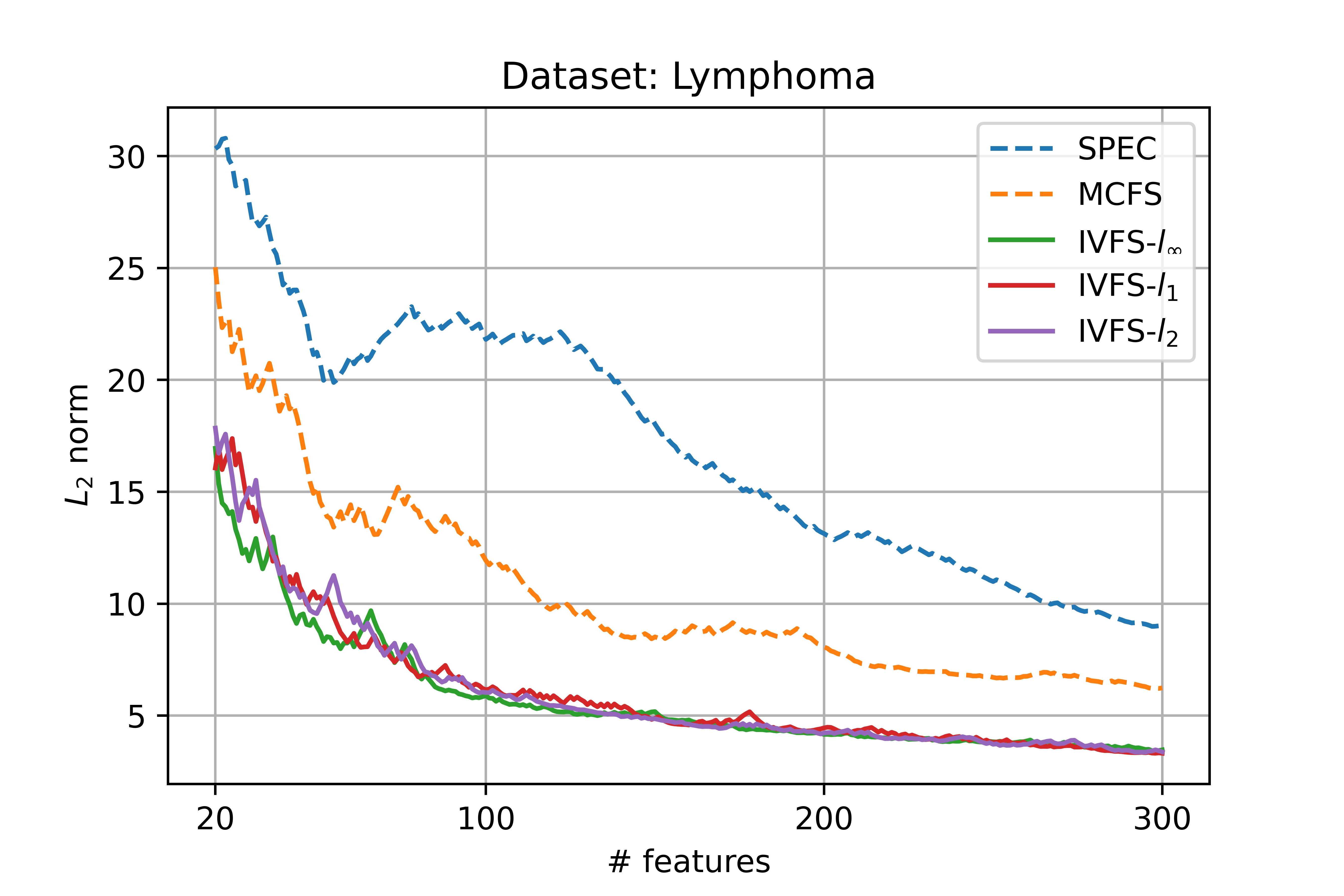}
	\end{center}
	\caption{$L_2$ norm v.s. the number of features using Lymphoma dataset.}
	\label{fig:12}
\end{figure}

\begin{figure}[ht]
	\begin{center}
		\includegraphics[width=0.87\linewidth]{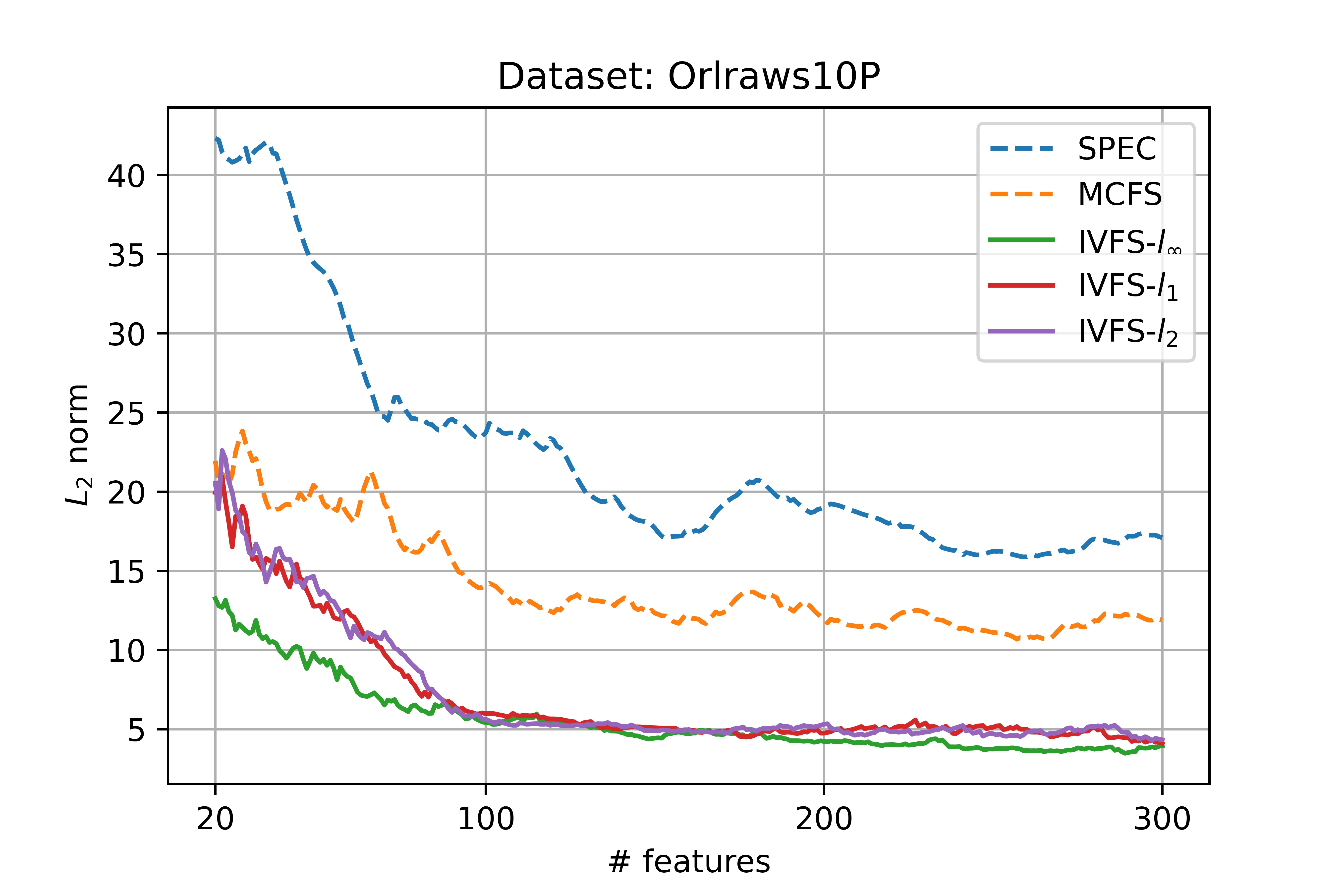}
	\end{center}
	\caption{$L_2$ norm v.s. the number of features using Orlraws10P dataset.}
	\label{fig:13}
\end{figure}

\begin{figure}[ht]
	\begin{center}
		\includegraphics[width=0.87\linewidth]{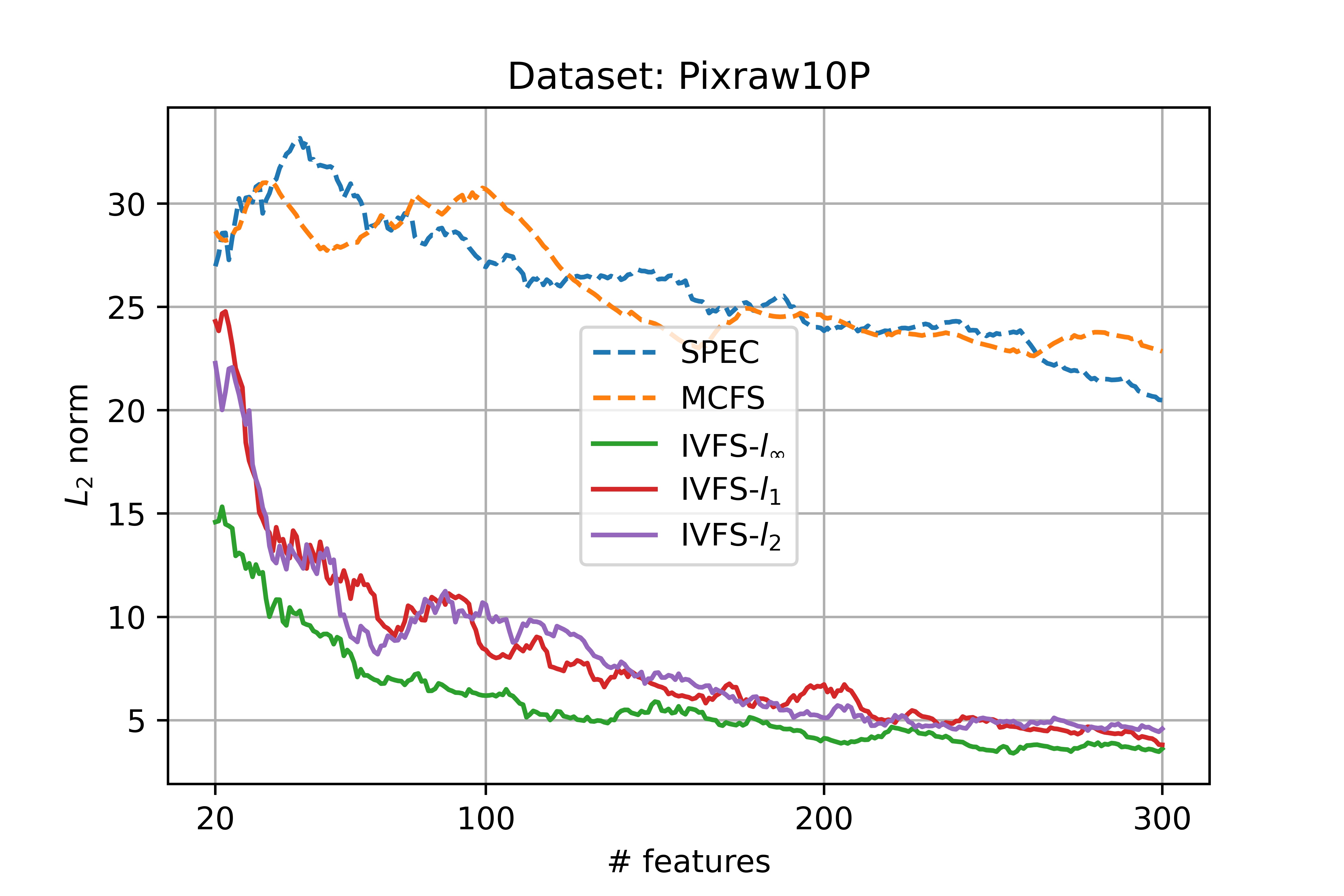}
	\end{center}
	\caption{$L_2$ norm v.s. the number of features using Pixraw10P dataset.}
	\label{fig:14}
\end{figure}

\begin{figure}[ht]
	\begin{center}
		\includegraphics[width=0.87\linewidth]{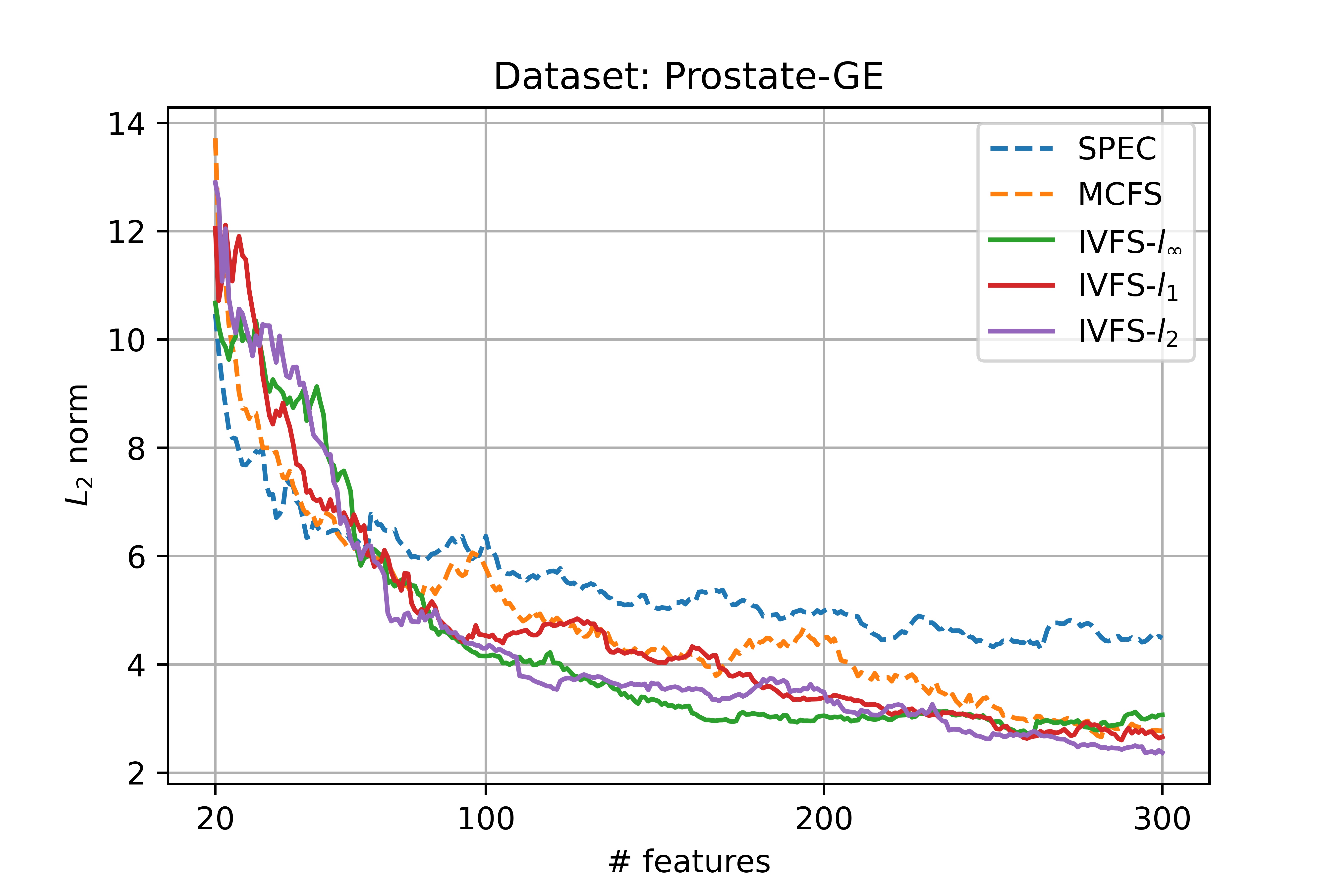}
	\end{center}
	\caption{$L_2$ norm v.s. the number of features using Prostate-GE dataset.}
	\label{fig:15}
\end{figure}

\begin{figure}[ht]
	\begin{center}
		\includegraphics[width=0.87\linewidth]{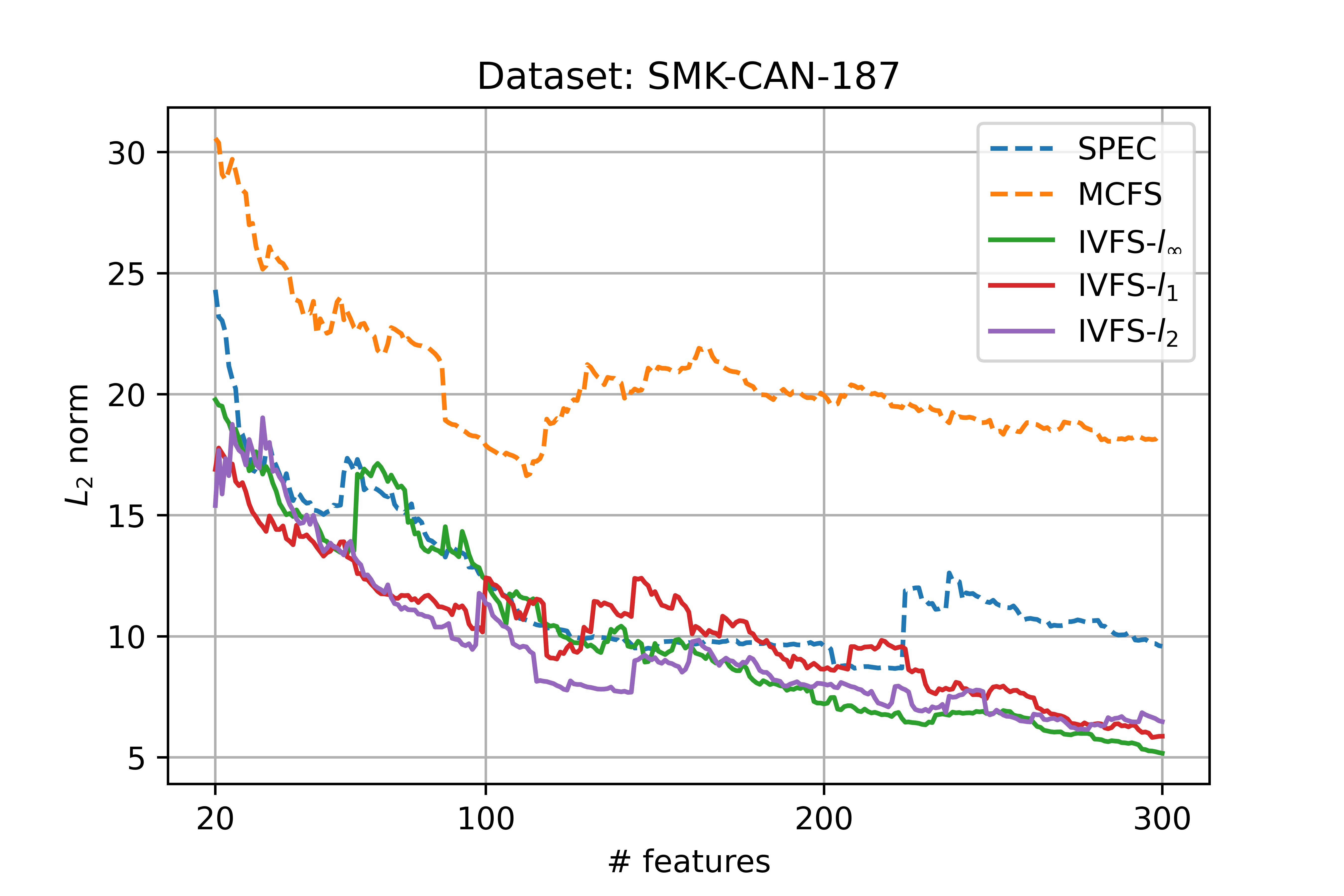}
	\end{center}
	\caption{$L_2$ norm v.s. the number of features using SMK-CAN-187 dataset.}
	\label{fig:16}
\end{figure}

\begin{figure}[ht]
	\begin{center}
		\includegraphics[width=0.87\linewidth]{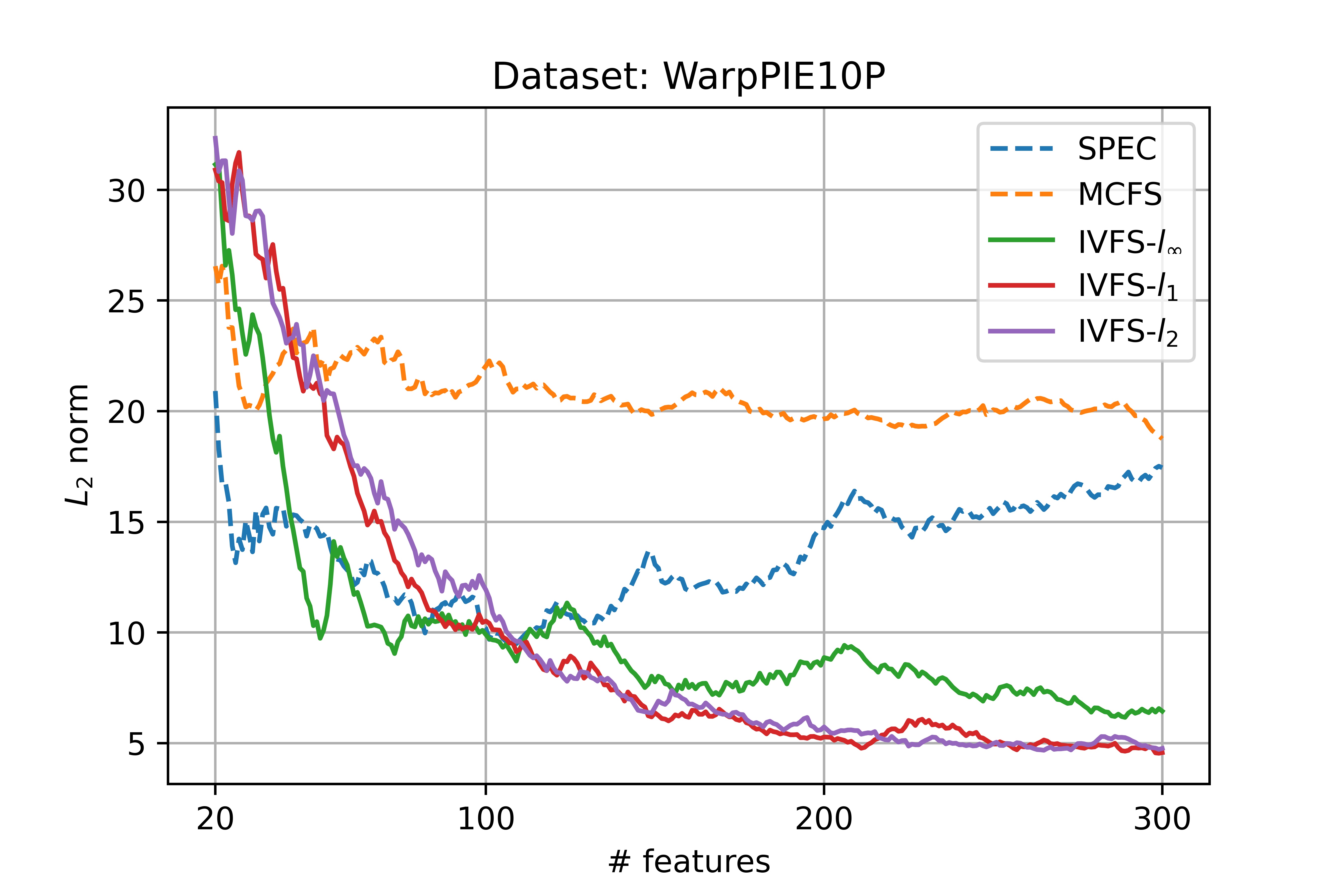}
	\end{center}
	\caption{$L_2$ norm v.s. the number of features using WarpPIE10P dataset.}
	\label{fig:17}
\end{figure}

\begin{figure}[ht]
	\begin{center}
		\includegraphics[width=0.87\linewidth]{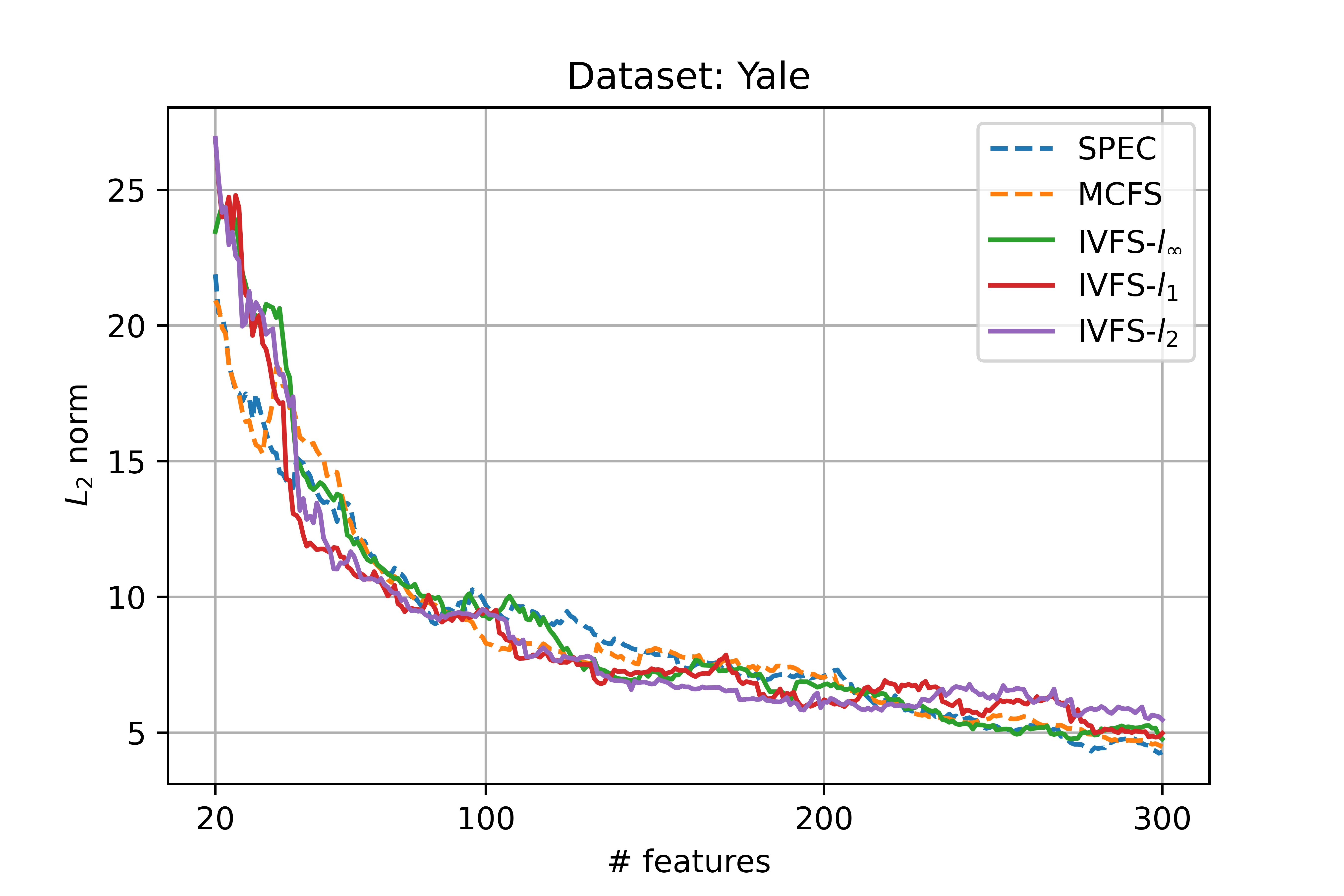}
	\end{center}
	\caption{$L_2$ norm v.s. the number of features using Yale dataset.}
	\label{fig:30}
\end{figure}

\begin{figure}[ht]
	\begin{center}
		\includegraphics[width=0.87\linewidth]{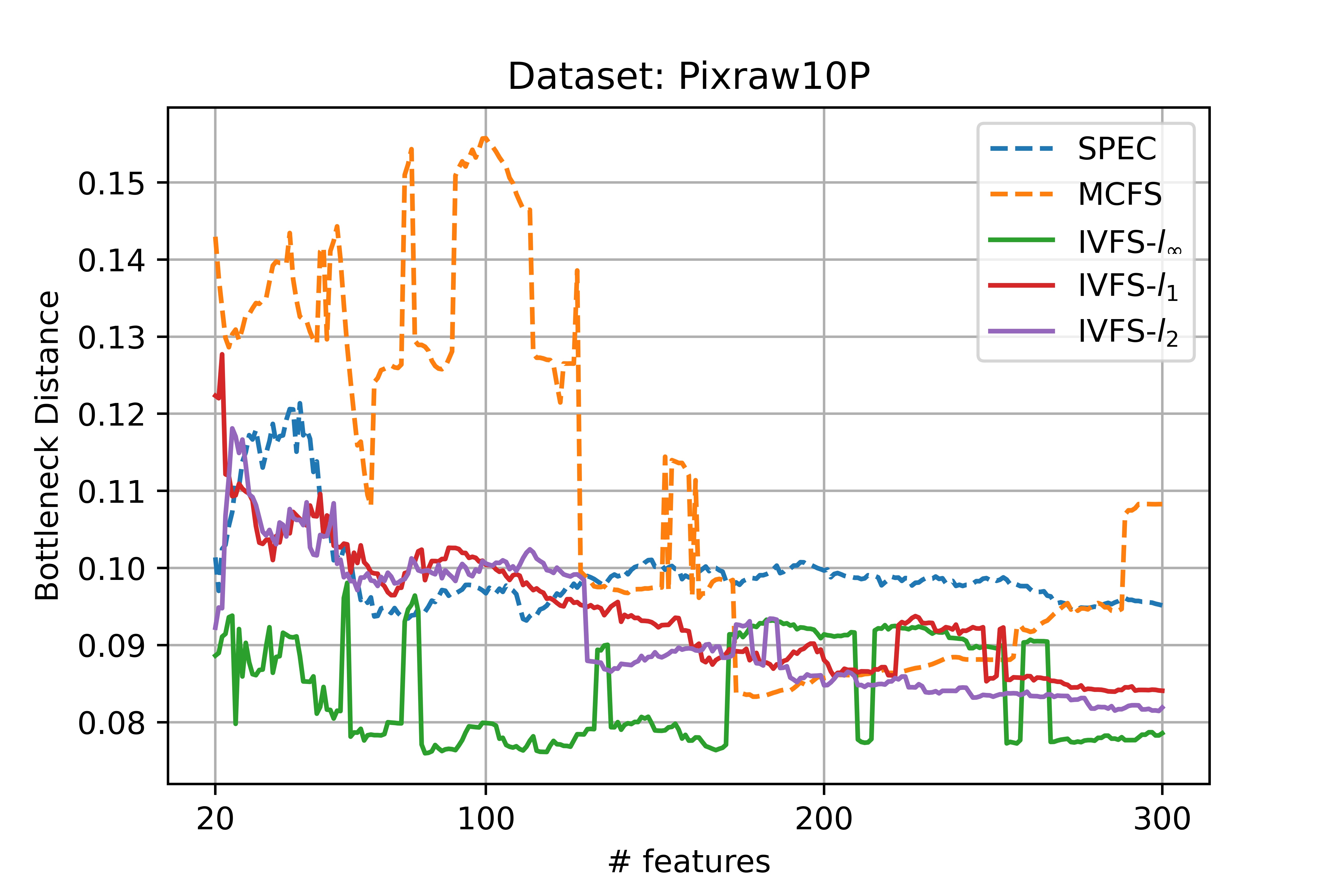}
	\end{center}
	\caption{Bottleneck distance v.s. the number of features using Pixraw10P dataset.}
	\label{fig:18}
\end{figure}

\begin{figure}[ht]
	\begin{center}
		\includegraphics[width=0.87\linewidth]{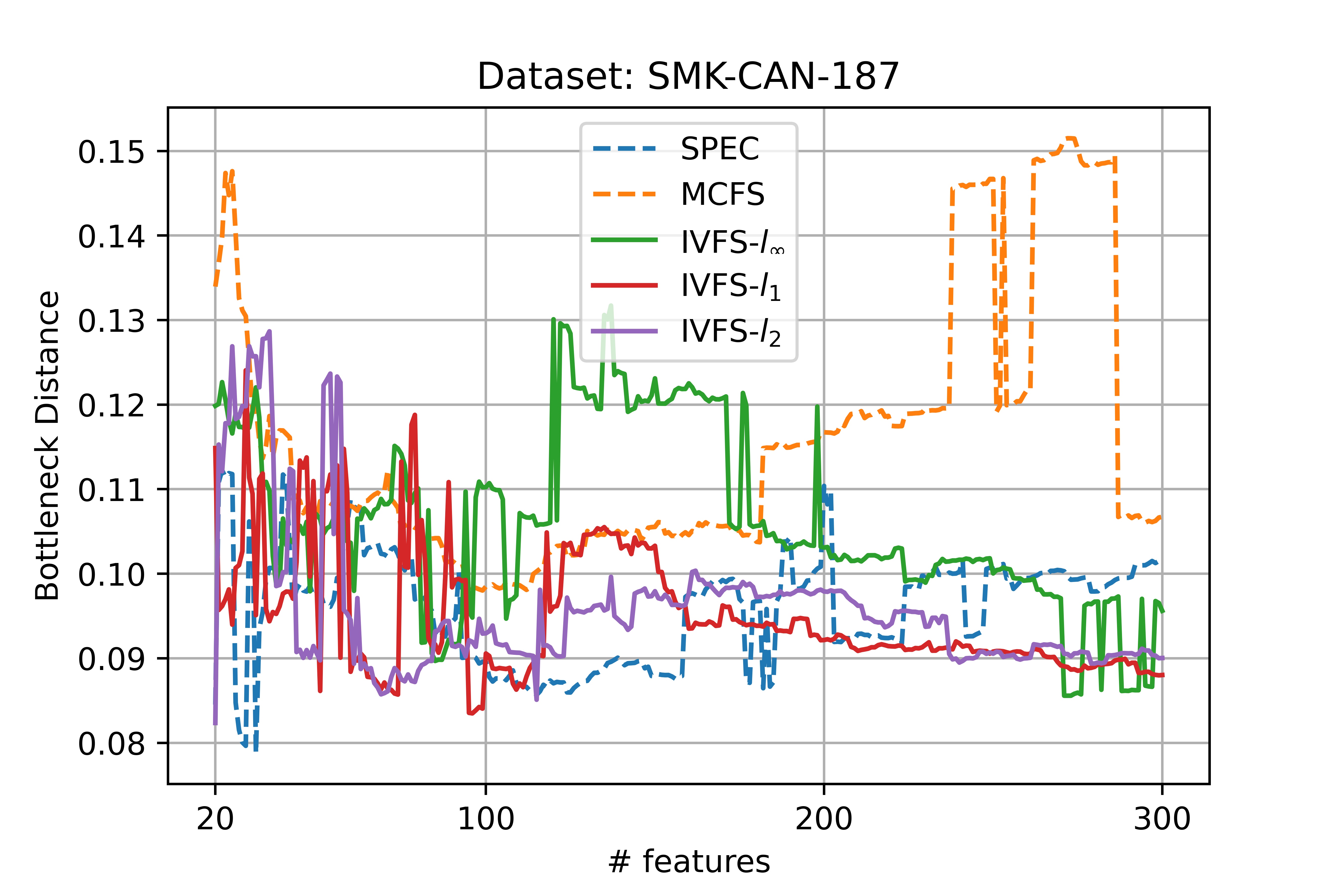}
	\end{center}
	\caption{Bottleneck distance v.s. the number of features using SMK-CAN-187 dataset.}
	\label{fig:19}
\end{figure}

Similar to the original paper's practice, We plot $\Vert D-D_F\Vert$ and $\omega_{\infty}(D-D_F)$ against the number of selected features from Figure~\ref{fig:12} to Figure~\ref{fig:19}, respectively. For $\Vert D-D_F\Vert$, we observe that IVFS performs better than the other two algorithms, and there is a clear trend that IVFS keeps lifting its performance as the number of features increases. This robustness comes from the fact that the inclusion value intrinsically contains rich information about features interactions. But for $\omega_{\infty}(D-D_F)$, although IVFS also performs well, the change of $\omega_{\infty}(D-D_F)$ with the increase of the number of features is very oscillatory and has no obvious regularity, as shown in Figure~\ref{fig:18} and Figure~\ref{fig:19}. In the original paper, these two figures are very smooth, because the original paper took the step of increasing \# features as 10, and we take it as 1. 

\subsection{Stability}
In principle, the selected feature pool should not vary significantly if we only change a few samples (by bootstrap), since the ``truly'' important features should be independent of the samples. We bootstrap samples from original dataset to mimic the process of sampling from population. We use same parameters as for testing the running time. In the discussion of stability, the original paper did not give the specific proportion of bootstrap data to original data, so we set it equal to 0.8. The results are averaged over 5 repetitions. In this way, we get Table~\ref{tab:2}. Here we use IVFS-$l_{\infty}$ as a representative of IVFS family.

\begin{figure}[ht]
	\begin{center}
		\includegraphics[width=1\linewidth]{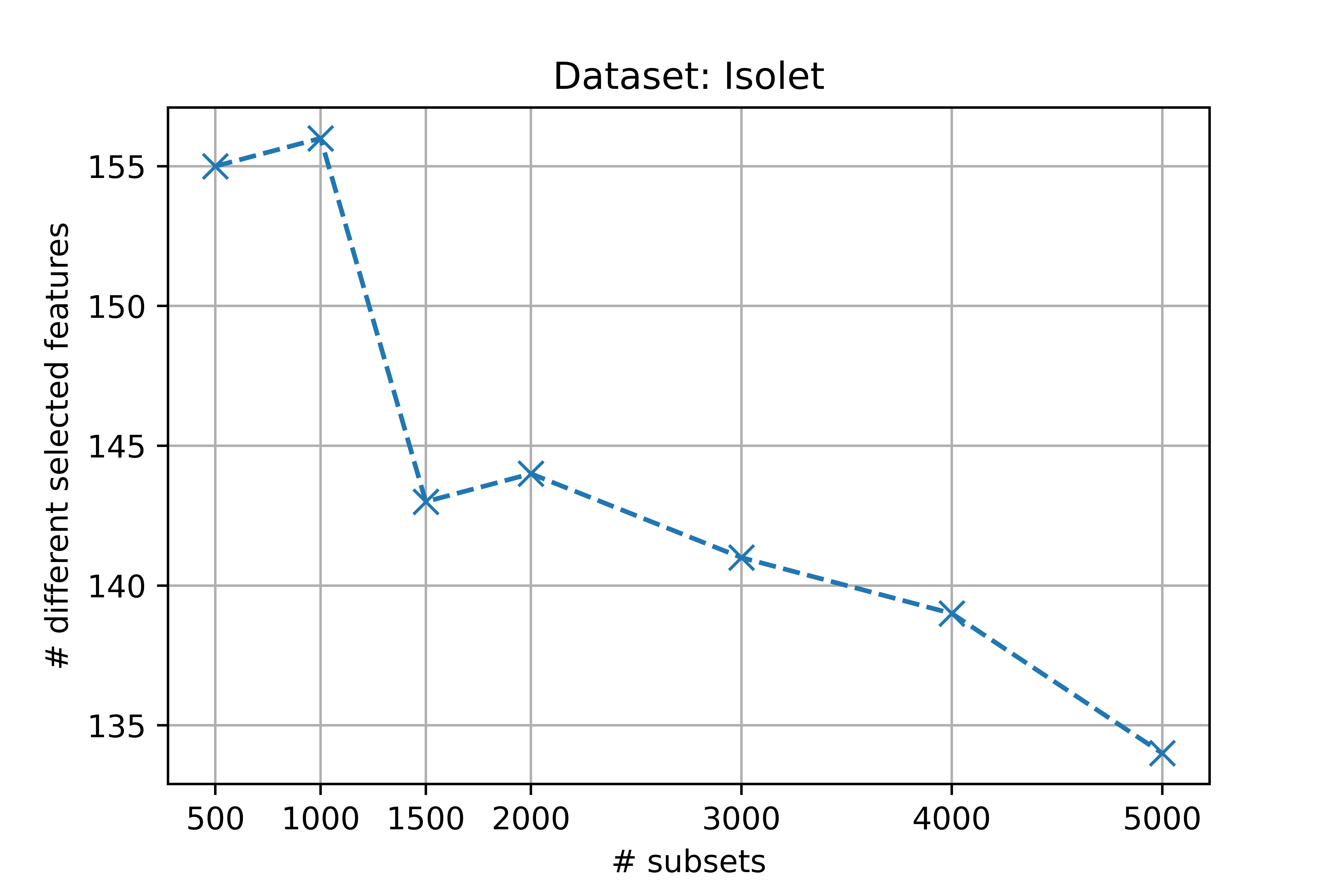}
	\end{center}
	\caption{The number of different selected features v.s. the number of subsets $k$ using Isolet dataset.}
	\label{fig:20}
\end{figure}

\begin{figure}[ht]
	\begin{center}
		\includegraphics[width=1\linewidth]{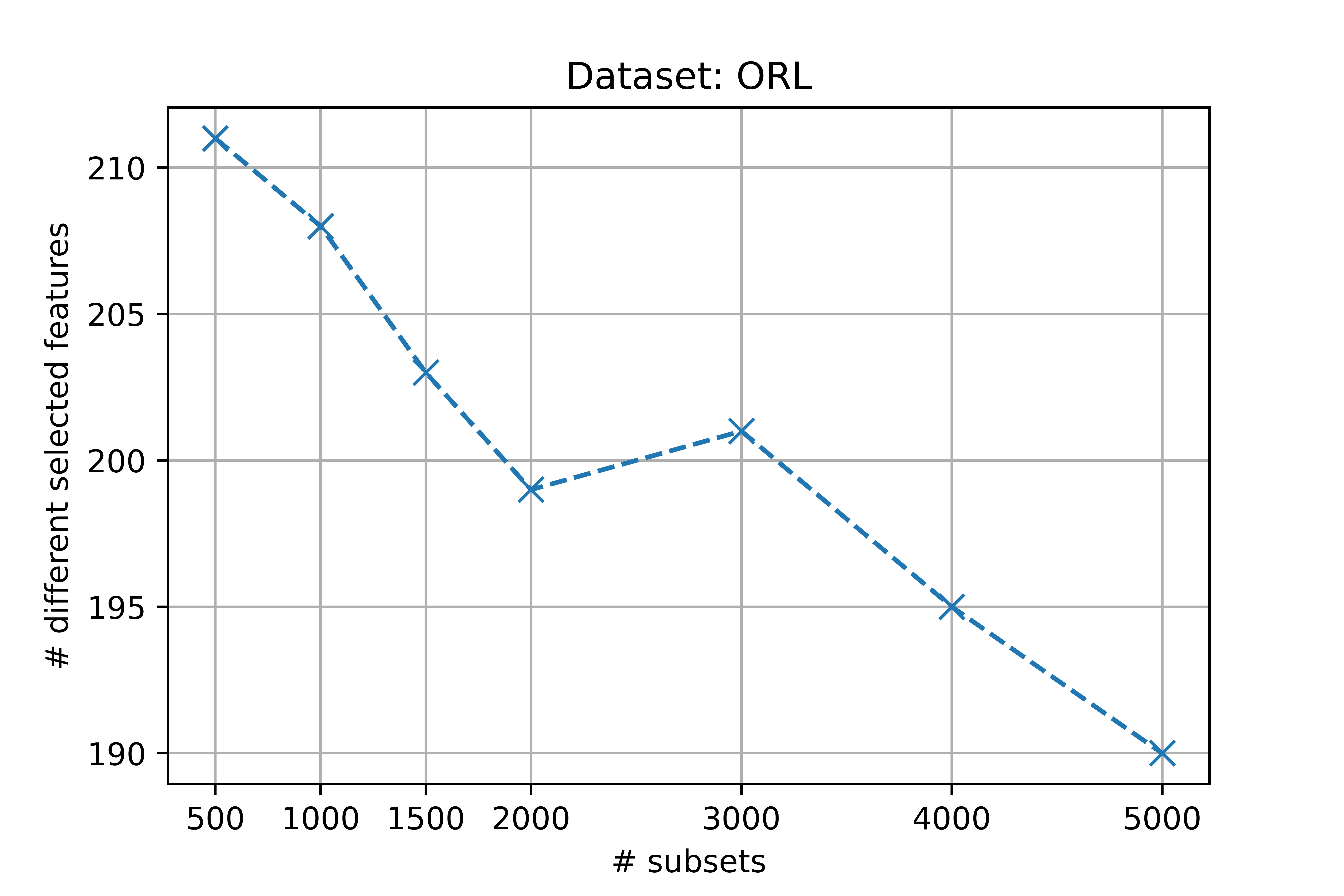}
	\end{center}
	\caption{The number of different selected features v.s. the number of subsets $k$ using ORL dataset.}
	\label{fig:21}
\end{figure}

\begin{table}[ht]
	\caption{Stability under bootstrap: the number of different selected features between original data and bootstrap data.}
	\begin{center}
		\begin{tabular}{c|ccc}
			\hline
			& SPEC & MCFS & IVFS-$l_{\infty}$ \\
			\hline
			Arcene & 18 & 87 & 292 \\
			CLL-SUB-111 & 0 & 230 & 289 \\
			COIL20 & 180 & 212 & 208 \\
			Isolet & 100 & 152 & 144 \\
			Lymphoma & 0 & 212 & 271 \\
			ORL & 200 & 214 & 204 \\
			Orlraws10P & 0 & 299 & 292 \\
			Pixraw10P & 285 & 294 & 289 \\
			Prostate-GE & 2 & 262 & 286 \\
			RELATHE & 278 & 276 & 280 \\
			SMK-CAN-187 & 0 & 96 & 294 \\
			WarpPIE10P & 266 & 220 & 258 \\
			Yale & 192 & 212 & 206\\
			
			\hline
		\end{tabular}
	\end{center}
	
	\label{tab:2}
\end{table}

We see that IVFS-$l_{\infty}$ is easily affected by the bootstrapping process. We think this is because the algorithm does not converge. So we consider changing the number of subsets $k$ to compare the number of different selected features. We randomly select two datasets, and the results are shown in the Figure~\ref{fig:20} and Figure~\ref{fig:21}. We can see that when we increase $k$, the number of different selected features shows a downward trend. This also shows a fatal drawback of the IVFS algorithm: if we want to get a convergent feature selection, we need to choose a very large value of $k$, which will cause the algorithm to run too long. 

\subsection{Efficiency-capacity trade-off}
From Table~\ref{tab:1}, we can see IVFS-$l_{\infty}$ is more efficient than SPEC and MCFS. In terms of running time, SPEC is the fastest and MCFS is the slowest. In particular, the convergence rate of MCFS in high-dimensional data sets is very slow, so it is not suitable for feature selection of high-dimensional data in practice. IVFS not only has good effect, but also has acceptable running time. 

\begin{figure}[ht]
	\begin{center}
		\includegraphics[width=1\linewidth]{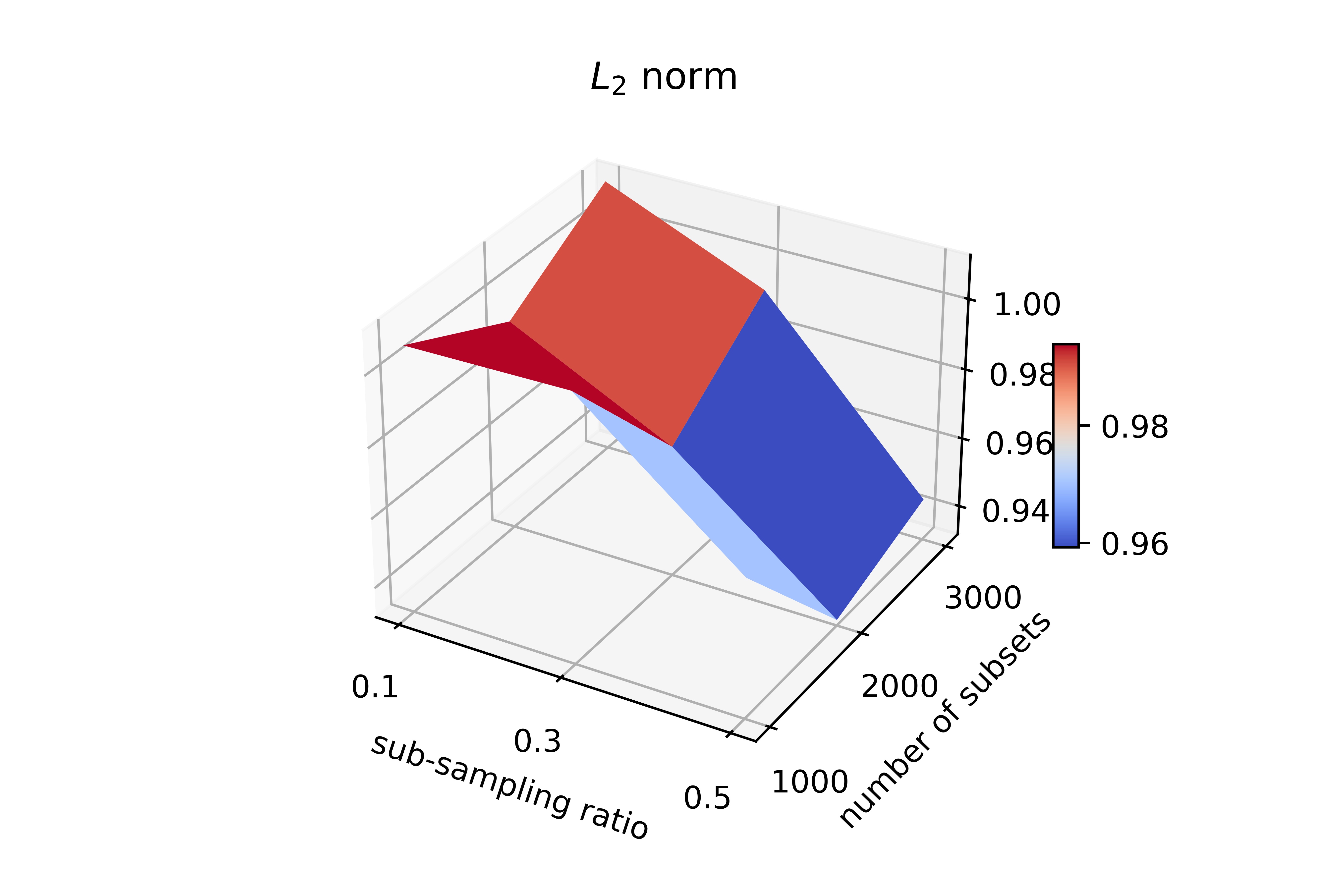}
	\end{center}
	\caption{$L_2$ norm performance comparison across different number of subsets $k$ and sub-sampling ratio $\tilde{n} / n$.}
	\label{fig:22}
\end{figure}

\begin{figure}[ht]
	\begin{center}
		\includegraphics[width=1\linewidth]{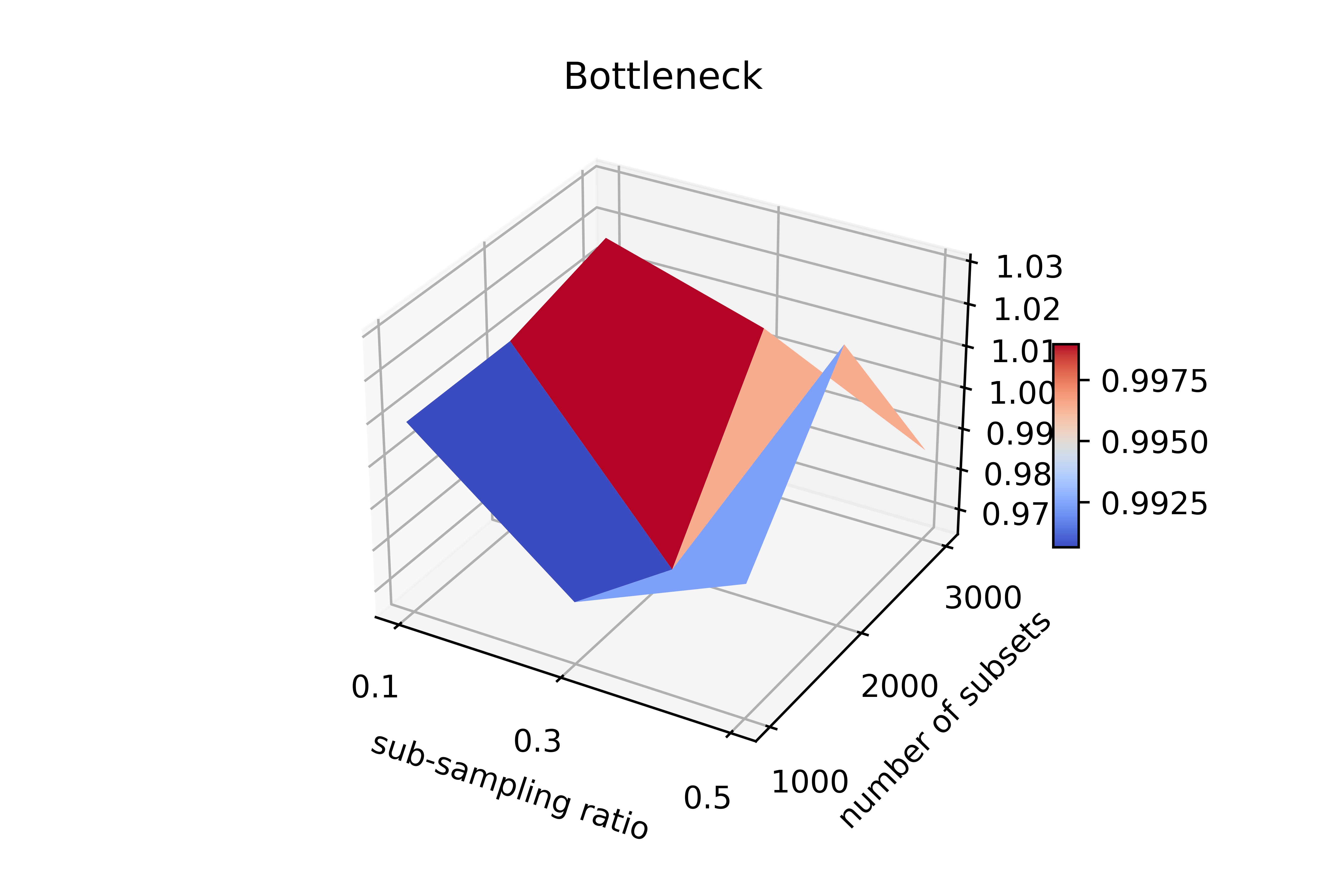}
	\end{center}
	\caption{Bottleneck distance performance comparison across different number of subsets $k$ and sub-sampling ratio $\tilde{n} / n$.}
	\label{fig:23}
\end{figure}

The computational cost of our IVFS algorithms depends tightly on the sub-sampling rate $\tilde{n} / n$ and the number of random subsets $k$. As one would expect, there exists a trade-off between computational efficiency and distance preserving power. Unlike the original paper, here we plot the relative performance (set the value for $k$ = 1000, $\tilde{n}$ = 0.1$n$ as 1) of different $k$ and $\tilde{n}$. The results are shown in the Figure~\ref{fig:22} and Figure~\ref{fig:23}. In general, $L_2$ norm performance of IVFS boosts as $k$ and $\tilde{n}$ increase because of more accurate estimate of the inclusion values. 

However, bottleneck distance performance has no obvious regularity. This may be related to the volatility of bottleneck distance. The results here are slightly different from those of the paper authors. This is not hard to understand. The third-party packages and parameters used in calculation of bottleneck distance are different from those of original authors. 

\section{Conclusion}
\label{conclusions}
In this paper, we implement the IVFS algorithm using our own code and reproduce the experimental results with datasets to compare and evaluate its performance against two other algorithms, SPEC and MCFS.

In the first half of this paper, we systematically explore the mathematical concepts underlying the IVFS algorithm, covering topics from similarity preservation to topology preservation, and from inclusion value to bottleneck distance. This section largely follows the original paper's framework. In the second half, we conduct a series of numerical experiments. Our results closely align with those reported in the original paper, indicating that our reproduction was successful. The findings demonstrate that IVFS is more effective than the other two methods. Additionally, the IVFS algorithm has a reasonable running time and can efficiently select features within a practical timeframe. IVFS performs exceptionally well with high-dimensional data, which aligns with the primary goal of feature selection.

However, some of our results differ from those in the original paper, with the most notable being the convergence behavior of IVFS. Specifically, when the number of subsets is relatively small, there is a significant discrepancy between the features selected from the original data and those from bootstrap samples. Additionally, the variation of bottleneck distance with the number of target features is oscillatory, unlike the smoother behavior seen with the $l_2$ norm. The mathematical foundation of the IVFS algorithm is commendable, as it uses topological space distances to measure similarity, allowing the selected features to preserve the original structure at a fundamental level. Although Theorem \ref{the:3} suggests that IVFS will eventually converge, the actual convergence rate is quite slow, at $O(1/\sqrt{k})$. Nevertheless, IVFS represents a unified feature selection scheme, and we anticipate that future variants will address these issues.

Our code is developed independently. To the best of our knowledge, the original authors did not release their source code, and all evaluation metrics are reproduced based on the descriptions in the original paper, which may result in differences.

\bibliographystyle{apalike}
\bibliography{egbib}

\newgeometry{left=1cm,right=1cm,top=1.5cm}
\renewcommand{\arraystretch}{1.4}
\begin{table*}[ht]
	\caption{Numerical experimental results. The unit of $L_1/n^2$ is ($\times10^{-2}$).}
	\newcommand{\tabincell}[2]{\begin{tabular}{@{}#1@{}}#2\end{tabular}}
	\begin{center}
		\begin{tabular}{c|ccc|c|ccccccccc}
			\hline
			$\mathbf{Dataset}$ & $\mathbf{n}$ & $\mathbf{d}$ & $\mathbf{C}$ & $\mathbf{Methods}$ & $\mathbf{KNN}$ & $\mathbf{NMI}$ & $\mathbf{K-means}$ & $\mathbf{\omega_1^1}$ & $\mathbf{\omega_{\infty}}$ & $\mathbf{L_{\infty}}$ & $\mathbf{L_1/n^2}$ & $\mathbf{L_2}$ & $\mathbf{Time(s)}$ \\
			
			\hline
			Arcene & 200 & \small{10000} & 2 
			& \tabincell{c}{SPEC \\ MCFS \\ IVFS-$l_{\infty}$} 
			& \tabincell{c}{72.5\% \\ $\mathbf{85.0\%}$ \\ $\mathbf{85.0\%}$ } 
			& \tabincell{c}{0.08 \\ 0.08 \\ $\mathbf{0.09}$ } 
			& \tabincell{c}{64.9\% \\ $\mathbf{65.0\%}$ \\ $\mathbf{65.0\%}$ } 
			& \tabincell{c}{0.19 \\ 0.20 \\ $\mathbf{0.16}$ } 
			& \tabincell{c}{0.06 \\ 0.06 \\ $\mathbf{0.05}$ }
			& \tabincell{c}{0.15 \\ $\mathbf{0.13}$ \\ $\mathbf{0.13}$ }
			& \tabincell{c}{3.15 \\ 2.65 \\ $\mathbf{2.63}$ }
			& \tabincell{c}{11.10 \\ 9.40 \\ $\mathbf{9.31}$ }
			& \tabincell{c}{2.13 \\ 196.5 \\ 10.64 } \\
			
			\hline
			\tabincell{c}{CLL-SUB\\-111} & 111 & \small{11340} & 3 
			& \tabincell{c}{SPEC \\ MCFS \\ IVFS-$l_{\infty}$} 
			& \tabincell{c}{69.6\% \\ 82.6\% \\ $\mathbf{87.0\%}$ } 
			& \tabincell{c}{0.21 \\ 0.25 \\ $\mathbf{0.27}$ } 
			& \tabincell{c}{53.3\% \\ 54.1\% \\ $\mathbf{56.9\%}$ } 
			& \tabincell{c}{0.00 \\ 0.00 \\ 0.00 } 
			& \tabincell{c}{0.20 \\ 0.09 \\ $\mathbf{0.06}$ }
			& \tabincell{c}{0.16 \\ $\mathbf{0.12}$ \\ 0.16 }
			& \tabincell{c}{3.00 \\ 3.06 \\ $\mathbf{2.87}$ }
			& \tabincell{c}{6.10 \\ $\mathbf{6.00}$ \\ 6.02 }
			& \tabincell{c}{2.02 \\ 162.3 \\ 6.72 } \\
			
			\hline
			COIL20 & \small{1440} & 1024  & 20 
			& \tabincell{c}{SPEC \\ MCFS \\ IVFS-$l_{\infty}$} 
			& \tabincell{c}{98.6\% \\ $\mathbf{99.3\%}$ \\ 98.6\% } 
			& \tabincell{c}{0.74 \\ $\mathbf{0.77}$ \\ 0.76 } 
			& \tabincell{c}{61.5\% \\ $\mathbf{64.4\%}$ \\ 63.3\% } 
			& \tabincell{c}{0.48 \\ 0.48 \\ $\mathbf{0.41}$ } 
			& \tabincell{c}{0.47 \\ 0.47 \\ $\mathbf{0.44}$ }
			& \tabincell{c}{0.24 \\ $\mathbf{0.09}$ \\ $\mathbf{0.09}$ }
			& \tabincell{c}{4.36 \\ $\mathbf{1.55}$ \\ $\mathbf{1.55}$ }
			& \tabincell{c}{115.4 \\ $\mathbf{39.52}$ \\ 39.94 }
			& \tabincell{c}{10.79 \\ 26.97 \\ 17.48 } \\
			
			\hline
			Isolet & \small{1560} & 617  & 26 
			& \tabincell{c}{SPEC \\ MCFS \\ IVFS-$l_{\infty}$} 
			& \tabincell{c}{85.9\% \\ 84.6\% \\ $\mathbf{86.2\%}$ } 
			& \tabincell{c}{0.75 \\ 0.72 \\ $\mathbf{0.76}$ } 
			& \tabincell{c}{58.9\% \\ 55.8\% \\ $\mathbf{62.6\%}$ } 
			& \tabincell{c}{0.42 \\ 0.33 \\ $\mathbf{0.26}$ } 
			& \tabincell{c}{0.11 \\ 0.10 \\ $\mathbf{0.08}$ }
			& \tabincell{c}{0.11 \\ $\mathbf{0.10}$ \\ $\mathbf{0.10}$ }
			& \tabincell{c}{2.09 \\ 1.71 \\ $\mathbf{1.60}$ }
			& \tabincell{c}{57.22 \\ 47.06 \\ $\mathbf{44.13}$ }
			& \tabincell{c}{8.00 \\ 5.93 \\ 10.87 } \\
			
			\hline
			Lymphoma & 96 & 4026  & 9 
			& \tabincell{c}{SPEC \\ MCFS \\ IVFS-$l_{\infty}$} 
			& \tabincell{c}{85.0\% \\ $\mathbf{100\%}$ \\ 95.0\% } 
			& \tabincell{c}{0.69 \\ 0.54 \\ $\mathbf{0.71}$ } 
			& \tabincell{c}{63.3\% \\ 51.4\% \\ $\mathbf{61.2\%}$ } 
			& \tabincell{c}{0.36 \\ 0.35 \\ $\mathbf{0.26}$ } 
			& \tabincell{c}{$\mathbf{0.04}$ \\ $\mathbf{0.04}$ \\ $\mathbf{0.04}$ }
			& \tabincell{c}{0.23 \\ 0.12 \\ $\mathbf{0.09}$ }
			& \tabincell{c}{5.24 \\ 2.33 \\ $\mathbf{1.84}$ }
			& \tabincell{c}{8.99 \\ 4.03 \\ $\mathbf{3.16}$ }
			& \tabincell{c}{0.72 \\ 39.04 \\ 2.14 } \\
			
			\hline
			ORL & 400 & \small{1024}  & 40 
			& \tabincell{c}{SPEC \\ MCFS \\ IVFS-$l_{\infty}$} 
			& \tabincell{c}{88.6\% \\ 90.0\% \\ $\mathbf{96.2\%}$ } 
			& \tabincell{c}{0.76 \\ 0.78 \\ $\mathbf{0.79}$ } 
			& \tabincell{c}{57.1\% \\ $\mathbf{59.6\%}$ \\ 59.1\% } 
			& \tabincell{c}{1.07 \\ 0.93 \\ $\mathbf{0.86}$ } 
			& \tabincell{c}{0.13 \\ $\mathbf{0.12}$ \\ $\mathbf{0.12}$ }
			& \tabincell{c}{0.17 \\ 0.08 \\ $\mathbf{0.07}$ }
			& \tabincell{c}{4.40 \\ 1.30 \\ $\mathbf{1.20}$ }
			& \tabincell{c}{29.4 \\ 9.29 \\ $\mathbf{8.63}$ }
			& \tabincell{c}{0.96 \\ 16.90 \\ 2.14 } \\
			
			\hline
			Orlraws10P & 100 & \small{10304}  & 10 
			& \tabincell{c}{SPEC \\ MCFS \\ IVFS-$l_{\infty}$} 
			& \tabincell{c}{85.0\% \\ $\mathbf{95.0\%}$ \\ $\mathbf{95.0\%}$ } 
			& \tabincell{c}{0.73 \\ 0.81 \\ $\mathbf{0.85}$ } 
			& \tabincell{c}{69.4\% \\ 75.6\% \\ $\mathbf{79.2\%}$ } 
			& \tabincell{c}{0.64 \\ 0.39 \\ $\mathbf{0.30}$ } 
			& \tabincell{c}{0.13 \\ 0.11 \\ $\mathbf{0.10}$ }
			& \tabincell{c}{0.42 \\ 0.19 \\ $\mathbf{0.08}$ }
			& \tabincell{c}{9.83 \\ 4.48 \\ $\mathbf{1.69}$ }
			& \tabincell{c}{17.10 \\ 7.90 \\ $\mathbf{3.00}$ }
			& \tabincell{c}{1.82 \\ 135.9 \\ 5.34 } \\
			
			\hline
			Pixraw10P & 100 & \small{10000}  & 10 
			& \tabincell{c}{SPEC \\ MCFS \\ IVFS-$l_{\infty}$} 
			& \tabincell{c}{$\mathbf{100\%}$ \\ 95.0\% \\ $\mathbf{100\%}$ } 
			& \tabincell{c}{0.77 \\ 0.88 \\ $\mathbf{0.93}$ } 
			& \tabincell{c}{77.6\% \\ 78.5\% \\ $\mathbf{92.8\%}$ } 
			& \tabincell{c}{0.56 \\ 0.48 \\ $\mathbf{0.33}$ } 
			& \tabincell{c}{0.10 \\ 0.09 \\ $\mathbf{0.07}$ }
			& \tabincell{c}{0.39 \\ 0.52 \\ $\mathbf{0.08}$ }
			& \tabincell{c}{11.61 \\ 11.36 \\ $\mathbf{1.68}$ }
			& \tabincell{c}{20.49 \\ 20.06 \\ $\mathbf{2.96}$ }
			& \tabincell{c}{1.82 \\ 129.3 \\ 5.41 } \\
			
			\hline
			\tabincell{c}{Prostate\\-GE} & 102 & 5966  & 2 
			& \tabincell{c}{SPEC \\ MCFS \\ IVFS-$l_{\infty}$} 
			& \tabincell{c}{90.5\% \\ $\mathbf{100\%}$ \\ $\mathbf{100\%}$ } 
			& \tabincell{c}{0.03 \\ $\mathbf{0.08}$ \\ $\mathbf{0.08}$ } 
			& \tabincell{c}{59.5\% \\ 62.8\% \\ $\mathbf{63.1\%}$ } 
			& \tabincell{c}{0.97 \\ $\mathbf{0.81}$ \\ $\mathbf{0.81}$ } 
			& \tabincell{c}{$\mathbf{0.08}$ \\ $\mathbf{0.08}$ \\ $\mathbf{0.08}$ }
			& \tabincell{c}{0.12 \\ 0.07 \\ $\mathbf{0.06}$ }
			& \tabincell{c}{2.38 \\ 1.56 \\ $\mathbf{1.31}$ }
			& \tabincell{c}{4.49 \\ 2.78 \\ $\mathbf{2.43}$ }
			& \tabincell{c}{1.05 \\ 69.62 \\ 3.19 } \\
			
			\hline
			RELATHE & \small{1427} & 4322  & 2 
			& \tabincell{c}{SPEC \\ MCFS \\ IVFS-$l_{\infty}$} 
			& \tabincell{c}{69.2\% \\ $\mathbf{73.4\%}$ \\ $\mathbf{73.4\%}$ } 
			& \tabincell{c}{0.00 \\ 0.00 \\ 0.00 } 
			& \tabincell{c}{54.6\% \\ $\mathbf{54.7\%}$ \\ $\mathbf{54.7\%}$ } 
			& \tabincell{c}{1.36 \\ 1.50 \\ $\mathbf{0.38}$ } 
			& \tabincell{c}{0.25 \\ 0.27 \\ $\mathbf{0.11}$ }
			& \tabincell{c}{0.72 \\ 0.32 \\ $\mathbf{0.28}$ }
			& \tabincell{c}{3.86 \\ 1.82 \\ $\mathbf{1.97}$ }
			& \tabincell{c}{130.2 \\ $\mathbf{62.21}$ \\ 67.11 }
			& \tabincell{c}{38.79 \\ 834.0 \\ 71.39 } \\
			
			\hline
			\tabincell{c}{SMK-CAN\\-187} & 187 & \small{19993}  & 2 
			& \tabincell{c}{SPEC \\ MCFS \\ IVFS-$l_{\infty}$} 
			& \tabincell{c}{68.4\% \\ $\mathbf{76.3\%}$ \\ 73.7\% } 
			& \tabincell{c}{0.00 \\ $\mathbf{0.01}$ \\ $\mathbf{0.01}$ } 
			& \tabincell{c}{51.5\% \\ 56.6\% \\ $\mathbf{57.2\%}$ } 
			& \tabincell{c}{0.92 \\ 0.70 \\ $\mathbf{0.55}$ } 
			& \tabincell{c}{0.10 \\ 0.09 \\ $\mathbf{0.08}$ }
			& \tabincell{c}{0.18 \\ 0.12 \\ $\mathbf{0.08}$ }
			& \tabincell{c}{2.74 \\ 2.51 \\ $\mathbf{1.53}$ }
			& \tabincell{c}{9.59 \\ 8.47 \\ $\mathbf{5.17}$ }
			& \tabincell{c}{4.12 \\ 587.1 \\ 19.14 } \\
			
			\hline
			\tabincell{c}{Warp\\-PIE10P} & 210 & 2420  & 10 
			& \tabincell{c}{SPEC \\ MCFS \\ IVFS-$l_{\infty}$} 
			& \tabincell{c}{$\mathbf{100\%}$ \\ $\mathbf{100\%}$ \\ $\mathbf{100\%}$ } 
			& \tabincell{c}{$\mathbf{0.34}$ \\ 0.33 \\ 0.32 } 
			& \tabincell{c}{$\mathbf{31.6\%}$ \\ 30.1\% \\ 29.2\% } 
			& \tabincell{c}{1.52 \\ $\mathbf{1.02}$ \\ 1.07 } 
			& \tabincell{c}{0.40 \\ $\mathbf{0.31}$ \\ $\mathbf{0.31}$ }
			& \tabincell{c}{0.18 \\ 0.11 \\ $\mathbf{0.05}$ }
			& \tabincell{c}{5.14 \\ 2.11 \\ $\mathbf{1.06}$ }
			& \tabincell{c}{17.44 \\ 7.58 \\ $\mathbf{3.89}$ }
			& \tabincell{c}{0.53 \\ 3.43 \\ 2.28 } \\
			
			\hline
			Yale & 165 & 1024  & 15 
			& \tabincell{c}{SPEC \\ MCFS \\ IVFS-$l_{\infty}$} 
			& \tabincell{c}{75.8\% \\ 75.8\% \\ $\mathbf{78.8\%}$ } 
			& \tabincell{c}{$\mathbf{0.57}$ \\ $\mathbf{0.57}$ \\ 0.56 } 
			& \tabincell{c}{44.6\% \\ 47.1\% \\ $\mathbf{47.9\%}$ } 
			& \tabincell{c}{0.64 \\ 0.69 \\ $\mathbf{0.57}$ } 
			& \tabincell{c}{$\mathbf{0.09}$ \\ $\mathbf{0.09}$ \\ $\mathbf{0.09}$ }
			& \tabincell{c}{$\mathbf{0.07}$ \\ $\mathbf{0.07}$ \\ $\mathbf{0.07}$ }
			& \tabincell{c}{1.47 \\ 1.53 \\ $\mathbf{1.25}$ }
			& \tabincell{c}{4.29 \\ 4.48 \\ $\mathbf{3.70}$ }
			& \tabincell{c}{0.23 \\ 10.30 \\ 1.14 } \\
			
			\hline
		\end{tabular}
	\end{center}
	\label{tab:1}
\end{table*}

\end{document}